\documentclass[10pt,twocolumn,letterpaper]{article}
\usepackage{cvpr}              %
\usepackage[dvipsnames]{xcolor}

\definecolor{cvprblue}{rgb}{0.21,0.49,0.74}
\usepackage[pagebackref,breaklinks,colorlinks,citecolor=cvprblue]{hyperref}

\usepackage{times}
\usepackage{epsfig}
\usepackage{graphicx}
\usepackage{amsmath}
\usepackage{amssymb}

\usepackage{multirow}
\usepackage{bbm}
\usepackage{multirow}
\usepackage{algorithm}
\usepackage[noend]{algpseudocode}
\usepackage{tabularx}
\usepackage{makecell}
\usepackage{microtype}
\usepackage{booktabs}
\usepackage{textcomp}
\usepackage{colortbl}
\usepackage{url}
\usepackage{mathalfa}
\usepackage{graphicx}
\usepackage{graphbox}
\usepackage{array}
\usepackage{enumitem}
\usepackage{animate}
\usepackage{xspace} 

\definecolor{rowblue}{RGB}{220,230,240}
\definecolor{myorchid}{RGB}{150,10,30}
\definecolor{myblue}{RGB}{10,30,250}
\definecolor{mygreen}{RGB}{10,190,10}
\definecolor{myred}{RGB}{190,20,20}
\definecolor{mypurple}{RGB}{255,100,255}

\newcommand{\newtext}[1]{#1}

\AtBeginDocument{%
 \abovedisplayskip=6pt plus 3pt minus 5pt
 \abovedisplayshortskip=0pt plus 3pt
 \belowdisplayskip=6pt plus 3pt minus 5pt
 \belowdisplayshortskip=5pt plus 3pt minus 4pt
}

\makeatletter
\renewcommand{\paragraph}{%
  \@startsection{paragraph}{4}%
  {\z@}{0.4ex \@plus 1ex \@minus .1ex}{-1em}%
  {\normalfont\normalsize\bfseries}%
}
\makeatother

\makeatletter
\@namedef{ver@everyshi.sty}{} %
\makeatother

\usepackage{tikz} %
\usetikzlibrary{calc} %
\usetikzlibrary{tikzmark} %
\usetikzlibrary{spy} %
\usetikzlibrary{shapes.misc} %
\usepackage{pgfplots} %
\usepackage{pgfplotstable} %
\pgfplotsset{compat=newest} %

\newcommand{\MT}{\mathcal{F}}
\newcommand{\SMT}{\mathcal{S}}
\newcommand{\FLOW}{F}
\newcommand{\MS}{S}
\newcommand{\Encoder}{E}
\newcommand{\Decoder}{D}

\newcommand{\pix}{\mathbf{p}}

\newcommand{\MotionRep}{spectral volume\xspace}
\newcommand{\MotionReps}{spectral volumes\xspace}

\newcommand{\ModalBases}{image-space modal bases\xspace}

\usepackage{cuted}

\usepackage[font=small]{caption}
\usepackage[outline]{contour}

\usepackage[capitalize]{cleveref}
\crefname{section}{Sec.}{Secs.}
\Crefname{section}{Section}{Sections}
\Crefname{table}{Table}{Tables}
\crefname{table}{Tab.}{Tabs.}

\def\paperID{3783} %
\def\confName{CVPR}
\def\confYear{2024}

\begin{document}

\title{Generative Image Dynamics}

\author{
Zhengqi Li \qquad
Richard Tucker \qquad
Noah Snavely \qquad
Aleksander Holynski
\\[0.5em]
Google Research  \ \ \
\ \ \
}

\maketitle

\begin{strip}
\centering
\includegraphics[width=0.95\textwidth]{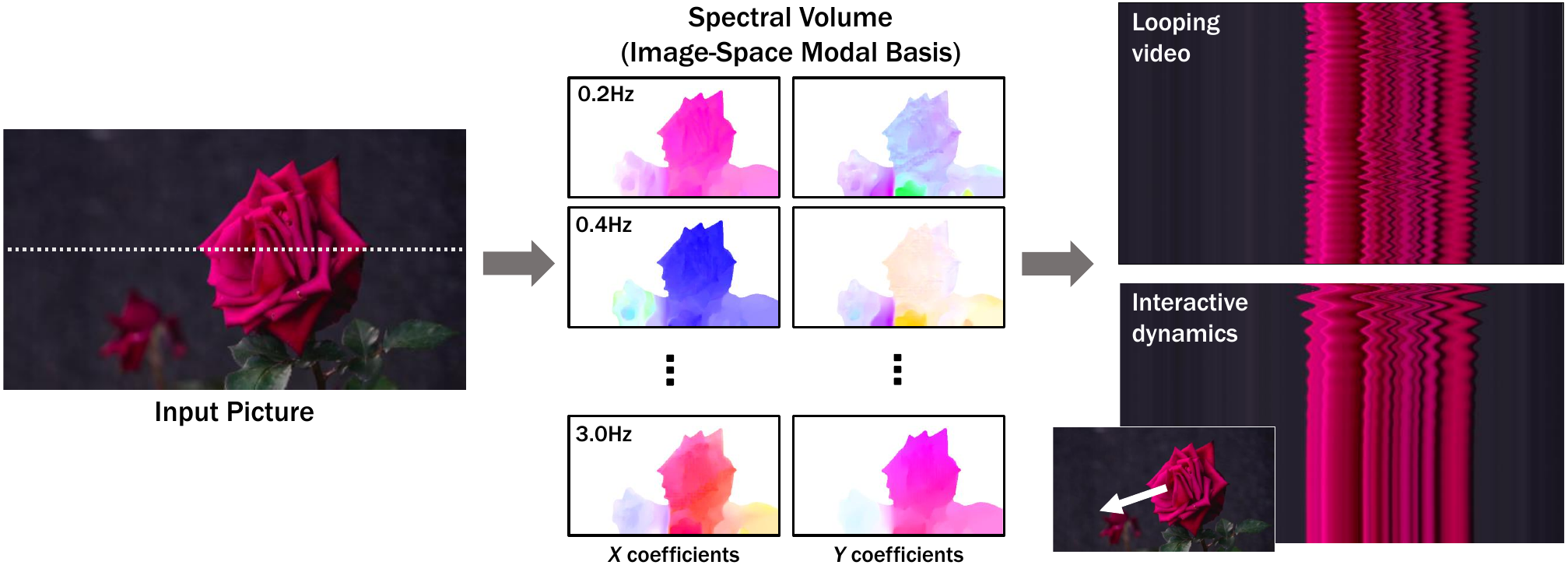} 
\vspace{-1em}
\captionof{figure}{
We model a generative image-space prior on scene motion: from a single RGB image, our method generates a \emph{spectral volume}~\cite{davis2016visual}, a motion representation that models dense, long-term pixel trajectories in the Fourier domain.
Our learned motion priors can be used to turn a single picture into a seamlessly looping video, or 
into an interactive simulation of dynamics that responds to user inputs like dragging and releasing points.
On the right, we visualize output videos as space-time $X$-$t$ slices (along the input scanline shown on the left). 
} \label{fig:teaser}
\end{strip}

\begin{abstract}
We present an approach to modeling an image-space prior on scene motion. 
Our prior is learned from a collection of motion trajectories extracted from real video sequences depicting natural, oscillatory dynamics of objects such as trees, flowers, candles, and clothes swaying in the wind. 
We model dense, long-term motion in the Fourier domain as \emph{\MotionReps}, which we find are well-suited to prediction with diffusion models.
Given a single image, our trained model uses a frequency-coordinated diffusion sampling process to predict a \MotionRep, which can be converted into a motion texture that spans an entire video.
Along with an image-based rendering module, the predicted motion representation can be used for a number of downstream applications, such as turning still images into seamlessly looping 
videos, or allowing users to interact with objects in real images, producing realistic simulated dynamics (by interpreting the \MotionReps as \ModalBases). See our project page for more results: \href{http://generative-dynamics.github.io/}{generative-dynamics.github.io}.

\end{abstract}

\section{Introduction}

The natural world 
is always in motion, with even seemingly static scenes containing subtle oscillations as a result of 
wind, water currents, respiration, or other natural rhythms. 
Emulating this motion is crucial in visual content synthesis---human sensitivity to motion can cause imagery without motion (or with slightly unrealistic motion) to
seem uncanny or unreal. 

While it is easy for humans to interpret or imagine motion in scenes, training a model to learn or produce realistic scene motion is far from trivial. The motion we observe in the world is the result of a scene's underlying physical dynamics, i.e., forces applied to objects that respond according to their unique physical properties---their mass, elasticity, etc---quantities that are hard to measure and capture at scale. Fortunately, measuring them is not necessary for certain applications:
\newtext{\eg, one can simulate plausible dynamics in a scene by simply analyzing some observed 2D motion~\cite{davis2016visual}.}

This same observed motion can also serve as a supervisory signal in learning dynamics \emph{across} scenes---because although observed motion is multi-modal and grounded in complex physical effects, it is nevertheless often predictable: candles will flicker in certain ways, trees will sway, and their leaves will rustle. 
For humans, this predictability is ingrained in our systems of perception: by viewing a still image, we can imagine plausible motions---
or, since there might have been many possible such motions, a \emph{distribution} of natural motions conditioned on that image. Given the facility with which humans are able to model these distributions, 
a natural research problem is to model them computationally.

Recent advances in generative models, in particular conditional diffusion models~\cite{sohl2015deep,song2020score, ho2020denoising}, have enabled us to model 
rich
distributions, including distributions of real images conditioned on text~\cite{saharia2022photorealistic, ramesh2022hierarchical, rombach2022high}. This capability has enabled 
several new 
applications, such as text-conditioned generation of 
diverse and realistic image content. Following the success of these image models, recent work has extended these models to
other domains, such as videos~\cite{ho2022imagen, blattmann2023align} and 3D geometry~\cite{watson2022novel, wynn-2023-diffusionerf, warburg2023nerfbusters, saxena2023surprising}.

In this paper, we 
model a generative prior for \emph{image-space scene motion}, i.e., the motion of all pixels in a single image. 
This model is trained on motion trajectories automatically extracted from a large collection of real video sequences. 
\newtext{In particular, from each training video we compute motion in the form of a \emph{spectral volume}~\cite{davis2015image, davis2016visual}, a frequency-domain representation of dense, long-range pixel trajectories. 
Spectral volumes are
well-suited to scenes that exhibit oscillatory dynamics, e.g., trees and flowers moving in the wind. 
We find that this representation is also highly
effective as an output of a diffusion model for modeling scene motion.
We train a generative model that, conditioned on a single image, can sample 
spectral volumes from its learned distribution.
A predicted spectral volume can then be directly transformed into a motion texture---a set of long-range, per-pixel 
motion trajectories---that can be used to animate the image. 
The spectral volume can also be interpreted as an \emph{image-space modal basis} 
for use in simulating
interactive dynamics~\cite{davis2015image}.
}

We predict spectral volumes from input images using a diffusion model that generates coefficients one 
frequency at a time, but coordinates these predictions across frequency bands through a shared attention module.
The predicted motions can be used to synthesize future frames (via an image-based rendering model)---turning still images into realistic animations, as illustrated in Fig.~\ref{fig:teaser}.

Compared with priors over raw RGB pixels, priors over motion capture more fundamental, lower-dimensional 
structure that efficiently explains long-range variations in pixel values. Hence, generating intermediate motion
leads to more coherent long-term generation and more fine-grained control over animations.
We demonstrate the use of our trained model in several downstream applications, such as creating seamless looping videos, editing the generated motions, and enabling interactive dynamic images via image-space modal bases, i.e., simulating the response of object dynamics to user-applied forces~\cite{davis2015image}.%

\section{Related Work}

\paragraph{Generative synthesis.}
Recent advances in generative models have enabled photorealistic synthesis of images conditioned on text prompts~\cite{dhariwal2021diffusion, saharia2022photorealistic, ramesh2022hierarchical, rombach2022high,chang2022maskgit, chang2023muse}. 
These 
text-to-image models can be augmented to synthesize video sequences by extending the generated image tensors along a time dimension \cite{ho2022imagen, skorokhodov2022stylegan, blattmann2023align, yu2023video, zhou2022magicvideo, luo2023videofusion, yu2023video, brooks2022generating}. 
While these methods 
can produce video sequences that capture the spatiotemporal statistics of real footage, 
these videos often suffer from artifacts 
like incoherent motion, unrealistic temporal variation in textures, and violations of physical constraints like preservation of mass.

\paragraph{Animating images.}
Instead of generating videos entirely from text, other techniques take as input a still picture and animate it. 
Many recent deep learning methods adopt a 3D-Unet architecture to produce video volumes directly~\cite{lee2018stochastic, he2023animate, guo2023animatediff, Dorkenwald_2021_CVPR, voleti2022MCVD, Hoppe2022DiffusionMF}. 
These models are effectively the same video generation models (but conditioned on image information instead of text), and exhibit similar artifacts to those mentioned above. 
One way to overcome these limitations is to not directly generate the video content itself, but instead animate an input source image through 
image-based rendering, i.e., moving the image content around according to motion derived from external sources such as a driving video~\cite{siarohin2019first, siarohin2021motion, siarohin2019animating, wang2022latent, karras2023dreampose}, motion or 3D geometry priors~\cite{walker2015dense, xue2019visualdynamics, weng2019photo, zhao2022thin, bowen2022dimensions, mallya2022implicit, ni2023conditional, holynski2021animating, mahapatra2021controllable, mahapatra2023synthesizing, endo2019animatinglandscape, watson2022novel, sugimoto2022water}, or user annotations~\cite{chuang2005animating, hao2018controllable, blattmann2021ipoke, franceschi2020stochastic, yin2023dragnuwa, wang2023videocomposer, zhang2023controlvideo, chen2023motion}. 
Animating images according to motion fields yields greater temporal coherence and realism, but these prior methods either require additional guidance signals or user input, or utilize limited motion representations.

\paragraph{Motion models and motion priors.}
\newtext{In computer graphics, natural, oscillatory 3D motion (e.g., water rippling or trees waving in the wind) can be 
modeled with noise that is shaped in the Fourier domain and then converted 
to time-domain motion fields~\cite{stam1996multiscale,shinya1992stochastic}.
Some of these methods rely on a modal analysis of the underlying dynamics of the system being simulated~\cite{stam1997stochastic,diener2009wind,davis2015image}.
These spectral techniques were adapted to animate plants, water, and clouds from single 2D pictures by Chuang \etal~\cite{chuang2005animating}, given user annotations. 
Our work is especially inspired by 
Davis~\cite{davis2016visual}, who 
connected modal analysis of a scene with the motions observed in a video of that scene, and 
used this analysis to simulate interactive dynamics from a video.
We adopt the frequency-space \emph{spectral volume} motion representation from Davis \etal, extract this representation from a large set of training videos, and show that spectral volumes are suitable for predicting motion from single images with diffusion models.}

Other methods have used various motion representations in \emph{prediction} tasks, where an image or video is used to inform a deterministic future motion estimate~\cite{pintea2014dejavu,gao2018im2flow}, or a more rich \emph{distribution} of possible motions~\cite{vondrick2016generating,walker2016uncertain,xue2019visualdynamics}.
\newtext{However, many of these methods predict an optical flow motion estimate (i.e., the instantaneous motion of each pixel), not full 
motion trajectories.
In addition, much of this prior work is focused on tasks like activity recognition, not on synthesis tasks. }
More recent work has demonstrated the advantages of modeling and predicting motion using generative models in a number of closed-domain settings such as humans and animals~\cite{tevet2022human, chen2023executing, du2023avatars, raab2023single, ahn2023can, zhang2023remodiffuse}.

\paragraph{Videos as textures.} Certain moving scenes can be thought of as a kind of texture---termed \emph{dynamic textures}~\cite{doretto2003dynamic}---that model videos as space-time samples of a stochastic process. Dynamic textures can represent smooth, natural motions like waves, flames, or moving trees, and have been widely used for video classification, segmentation or encoding~\cite{casas20144d, chan2007classifying, chan2008modeling, saisan2001dynamic, chan2005mixtures}.
A related kind of texture, called a \emph{video texture}, represents a moving scene as a set of input video frames along with transition probabilities between any pair of frames~\cite{schodl2000video, narasimhan2022strumming}. 
A number of methods estimate 
dynamic or video textures through analysis of scene motion and pixel statistics, with the aim of generating seamlessly looping or infinitely varying output videos~\cite{liao2013automated, liao2015fast, agarwala2005panoramic, schodl2000video, couture2013omnistereo, flagg2009human}. In contrast to much of this 
work, our method learns priors in advance that can then be applied to single images.

\section{Overview}
Given a single picture $I_0$, our goal is to generate a video $\{ \hat{I}_1, \hat{I}_2., ..., \hat{I}_T\}$ featuring oscillatory motions such as those of trees, flowers, or candle flames swaying in the breeze. 
Our system consists of two modules: a motion prediction module and an image-based rendering module.
Our pipeline begins by using a latent diffusion model (LDM) to predict a spectral volume $\SMT = \left( S_{f_0}, S_{f_1}, ..., S_{f_{K-1}} \right)$ for the input $I_0$. 
The predicted spectral volume is then transformed to a motion texture $\MT = \left( F_1, F_2, ..., F_{T} \right)$ through an inverse discrete Fourier transform. This motion determines the position of each input pixel at every future time step. 

Given a predicted motion texture, we then animate the input RGB image using a neural image-based rendering technique (Sec.~\ref{sec:rendering}).
We explore applications of this method, including producing seamless looping animations and simulating interactive dynamics, in Sec.~\ref{sec:app}.

\section{Predicting motion} \label{sec:nsmt}

\begin{figure}[tb]
    \centering
    \setlength{\tabcolsep}{0.00cm}
    \begin{tabular}{cc}
    \includegraphics[width=0.5\columnwidth]{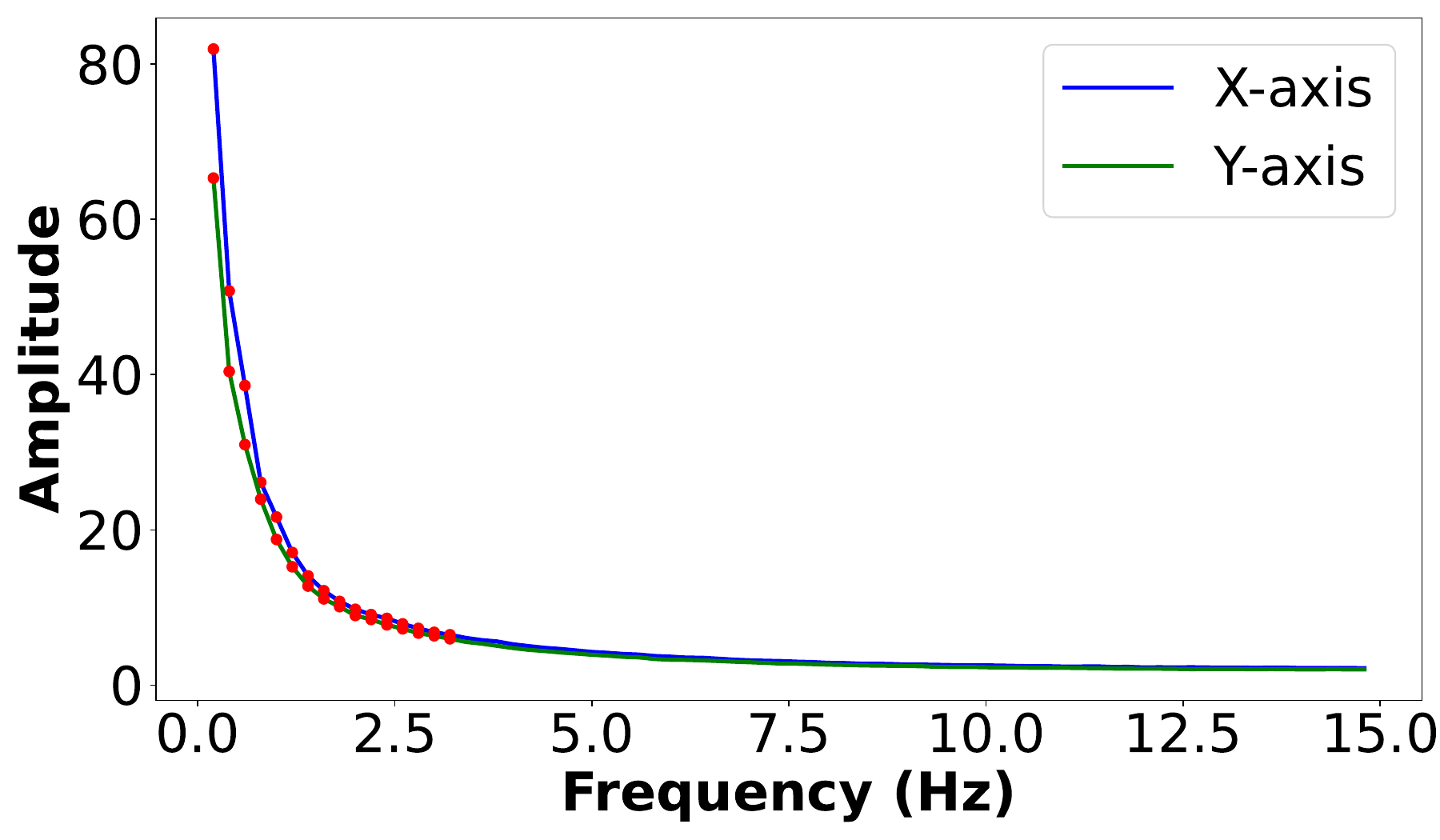} &
    \includegraphics[width=0.5\columnwidth]{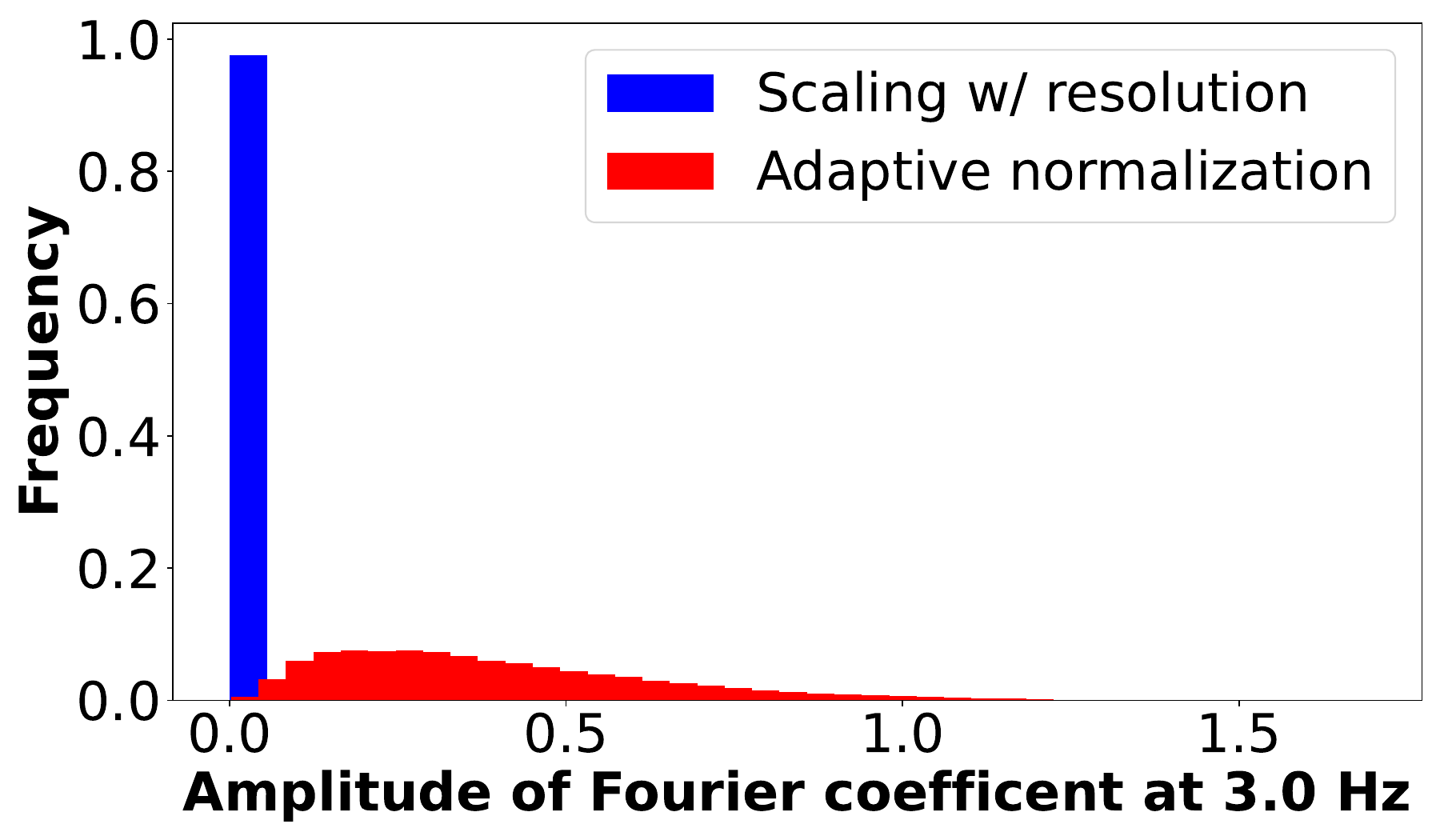} \\
    \vspace{-0.5cm}
    \end{tabular}
    \caption{\textbf{Left:} We visualize the average power spectrum for the $X$ and $Y$ motion components extracted from real videos, shown as the blue and green curves. Natural oscillation motions are composed primarily of low-frequency components, and so we use the first $K=16$ terms, marked with red dots. \textbf{Right:} we show a histogram of the amplitude of Fourier terms at $3.0$ Hz after (1) scaling amplitude by image width and height (blue), or (2) frequency adaptive normalization (red). Our adaptive normalization prevents the coefficients from concentrating at extreme values.}
    \label{fig:power_spectrum}
\end{figure}

\subsection{Motion representation}

\begin{figure*}[t]
\centering
  \includegraphics[width=0.95\textwidth]{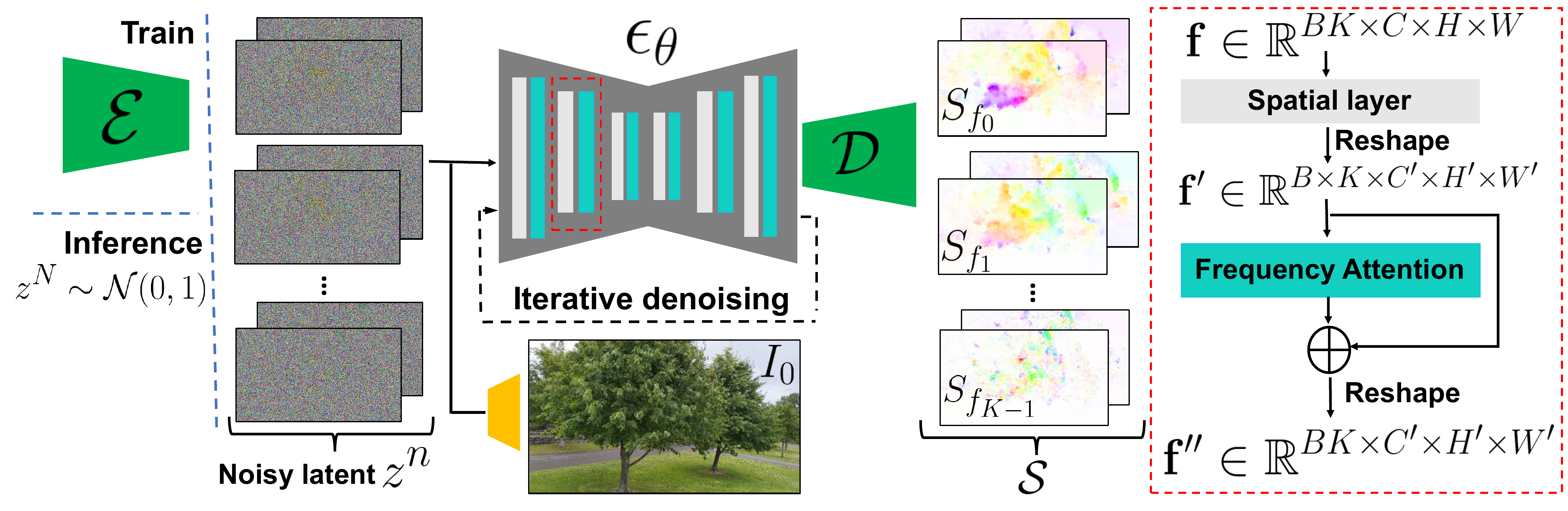} 
  \caption{\textbf{Motion prediction module.}
  We predict a spectral volume $\SMT$ through a frequency-coordinated denoising model. Each block of the diffusion network $\epsilon_{\theta}$ interleaves 2D spatial layers with attention layers (\textbf{red box, right}), and iteratively denoises latent features $z^n$. The denoised features are fed to a decoder $\mathcal{D}$ to produce $\SMT$. During training, we concatenate the downsampled input $I_0$ with noisy latent features encoded from a real motion texture via an encoder $\mathcal{E}$, and replace the noisy features with Gaussian noise $z^N$ during inference (\textbf{left}).
  }
\label{fig:overview_1}
\end{figure*}

Formally, a motion texture is a sequence of time-varying 2D displacement maps $\MT = \{ \FLOW_t | t = 1,..., T\}$, where the 2D displacement vector $\FLOW_t (\pix)$ at each pixel coordinate $\pix$ from input image $I_0$ defines the position of that pixel at a future time $t$~\cite{chuang2005animating}. 
To generate a future frame at time $t$, one can splat pixels from $I_0$ using the corresponding displacement map $D_t$, resulting in a forward-warped image $I'_{t}$:
\begin{align}
        I'_{t} (\pix + \FLOW_t(\pix)) = I_0 (\pix).
\end{align}
If our goal is to produce a video via a motion texture, then one choice would be to predict a time-domain motion texture directly from an input image.
However, the size of the motion texture would need to scale with the length of the video: generating $T$ output frames implies predicting $T$ displacement fields. 
To avoid predicting such a large output representation for long videos, many prior animation methods 
either generate video frames autoregressively~\cite{li2022infinitenature, liu2021infinite, endo2019animatinglandscape, voleti2022MCVD, blattmann2023align}, or predict each future output frame independently via an extra time embedding~\cite{bertiche2023blowing}.
However, neither strategy ensures long-term temporal consistency of generated videos.

Fortunately, many natural motions
can be described as a superposition of a small number of harmonic oscillators represented with different frequencies, amplitude and phases~\cite{chuang2005animating, diener2009wind, ota20031, kanda2003efficient, davis2016visual}. Because these underlying motions are quasi-periodic, it is natural to model them in the frequency domain.
Hence, we adopt from Davis~\etal~\cite{davis2016visual} an efficient frequency space representation of motion in a video called a \emph{spectral volume}, visualized in Fig.~\ref{fig:overview_1}.
A spectral volume is the temporal Fourier transform of per-pixel trajectories extracted from a video. 

Given this motion representation, we formulate the motion prediction problem as a multi-modal image-to-image translation task: from an input image to an output motion spectral volume.
We adopt latent diffusion models (LDMs) to generate spectral volumes comprised of
a $4K$-channel 2D motion spectrum map, 
where $K << T$ is the number of frequencies modeled, and where at each frequency we need four scalars to represent the complex Fourier coefficients for the $x$- and $y$-dimensions. 
Note that the motion trajectory of a pixel at future time steps 
$\MT (\pix) = \{ \FLOW_t (\pix) | t=1,2,...T \}$ 
and its representation as a spectral volume
$\SMT (\pix) = \{ \MS_{f_k}(\pix) | k = 0,1,..\frac{T}{2} - 1 \}$
are related by the Fast Fourier transform (FFT):
\begin{align}
    \SMT (\mathbf{p})= \text{FFT}(\MT (\pix)).
\end{align}

How should we select the $K$ output frequencies?
Prior work in real-time animation has observed that most natural oscillation motions are composed primarily of 
low-frequency components~\cite{diener2009wind, ota20031}. To validate this observation, we computed the average power spectrum of the motion extracted from 1,000 randomly sampled 5-second real video clips. 
As shown in the left plot of Fig.~\ref{fig:power_spectrum},  the power spectrum of the motion decreases exponentially with increasing frequency. This suggests that most natural oscillation motions can indeed be well represented by low-frequency terms. 
In practice, we found that the first $K=16$ Fourier coefficients are sufficient to realistically reproduce the original natural motion in a range of real videos and scenes.

\subsection{Predicting motion with a diffusion model}

We choose a latent diffusion model (LDM)~\cite{rombach2022high} as the backbone for our motion prediction module, as LDMs are more computationally efficient than pixel-space diffusion models, while preserving synthesis quality.
A standard LDM consists of two main modules: 
(1) a variational autoencoder (VAE) that compresses the input image to a latent space through an encoder $z = \Encoder(I)$, then reconstructs the input from the latent features via a decoder $I = \Decoder(z)$, and 
(2) a U-Net based diffusion model that learns to iteratively denoise features starting from Gaussian noise. 
Our training applies this process not to RGB images but to spectral volumes from real video sequences, which are encoded and then diffused for $n$ steps with a pre-defined variance schedule to produce noisy latents $z^{n}$. 
The 2D U-Nets are trained to denoise the noisy latents by iteratively estimating the noise $\epsilon_\theta (z^n; n, c)$ used to update the latent feature at each step $n \in (1, 2, ..., N)$. 
The training loss for the LDM is written as 
\begin{align}
    \mathcal{L}_\text{LDM} = \mathbb{E}_{n \in \mathcal{U}[1, N] , \epsilon^n \in \mathcal{N}(0, 1)} \left[|| \epsilon^n - \epsilon_{\theta} (z^n; n, c) ||^2 \right]
\end{align} %
where $c$ is the embedding of any conditional signal, such as text, or, in our case, the first frame of the training video sequence, $I_0$. The clean latent features $z^{0}$ are then passed through the decoder to recover the spectral volume.

\paragraph{Frequency adaptive normalization.}

One issue we observed is that motion textures have particular distribution characteristics across frequencies. As visualized in the left plot of Fig.~\ref{fig:power_spectrum}, the amplitude of the spectral volumes spans a range of 0 to 100 and decays approximately exponentially with increasing frequency. As diffusion models require that the absolute values of the output are between -1 and 1 for stable training and denoising~\cite{ho2020denoising}, we must normalize the coefficients of $\SMT$ extracted from real videos before using them for training.
If we scale the magnitudes of these coefficients to [0,1] based on the image dimensions as in prior work~\cite{endo2019animatinglandscape, saxena2023surprising}, 
nearly all the coefficients at higher frequencies will end up close to zero, as shown in the right plot of  Fig.~\ref{fig:power_spectrum}. 
Models trained on such data can produce inaccurate motions, since during inference, even small prediction errors can cause large relative errors after denormalization. %

To address this issue, we employ a simple but effective frequency adaptive normalization method: 
First, we independently normalize Fourier coefficients at each frequency based on statistics computed from the training set. Namely, for each individual frequency $f_j$, we compute the $95^\textrm{th}$ percentile of 
Fourier coefficient magnitudes over all input samples and use that value as a per-frequency scaling factor $s_{f_j}$.
We then apply a power transformation to each scaled Fourier coefficient to pull it away from extreme values.  In practice, we observe that a square root performs better than other nonlinear transformations such as log or reciprocal. In summary, the final coefficient values of spectral volume $\SMT(\pix)$ at frequency $f_j$ (used for training our LDM) are computed as
\begin{align}
    S'_{f_j}(\pix) = \text{sign}(S_{f_j})\sqrt{\left|\frac{S_{f_j}(\pix)}{s_{f_j}}\right|}.
\end{align}
As shown on the right plot of Fig.~\ref{fig:power_spectrum}, after applying frequency adaptive normalization, the spectral volume coefficients distribute more evenly.

\paragraph{Frequency-coordinated denoising.}

The straightforward way to predict a spectral volume $\SMT$ with $K$ frequency bands is to output a tensor of $4K$ channels from a single diffusion U-Net. However, as in prior work~\cite{blattmann2023align}, we observe that training a model to produce a large number of channels 
can yield over-smoothed, inaccurate outputs. An alternative would be to independently predict each individual frequency slice by injecting an extra frequency embedding into the LDM~\cite{bertiche2023blowing}, but this design choice would result in uncorrelated predictions in the frequency domain, leading to unrealistic motion.

Therefore, inspired by recent video diffusion work~\cite{blattmann2023align}, we propose a frequency-coordinated denoising strategy, illustrated in Fig.~\ref{fig:overview_1}. 
In particular, given an input image $I_0$, we first train an LDM $\epsilon_{\theta}$ to predict a single 4-channel frequency slice of spectral volume $S_{f_j}$, 
where we inject an extra frequency embedding along with the time-step embedding into the LDM.
We then freeze the parameters of this LDM
$\epsilon_{\theta}$, introduce attention layers interleaved with the 2D spatial layers of $\epsilon_{\theta}$ across the $K$ frequency bands, and fine-tune. Specifically, for a batch size $B$, the 2D spatial layers of $\epsilon_{\theta}$ treat the corresponding $B \cdot K$ noisy latent features of channel size $C$ as independent samples with shape $\mathcal{R}^{(B \cdot K) \times C \times H \times W}$. The attention layer then interprets these as consecutive features spanning the frequency axis, and we reshape the latent features from previous 2D spatial layers to $\mathcal{R}^{B \times K \times C \times H \times W}$ before feeding them to the attention layers. 
In other words, the frequency attention layers are fine-tuned to coordinate 
all frequency slices 
so as to produce coherent spectral volumes. 
In our experiments, we 
see that the average VAE reconstruction error improves from $0.024$ to $0.018$ when we switch from a single 2D U-Net to a frequency-coordinated denoising module, suggesting an improved upper bound on LDM prediction accuracy; in 
Sec.~\ref{sec:ablation}, we also 
show that this design choice improves video generation quality.

\section{Image-based rendering} \label{sec:rendering}

\begin{figure}[t]
\centering
  \includegraphics[width=1.0\columnwidth]{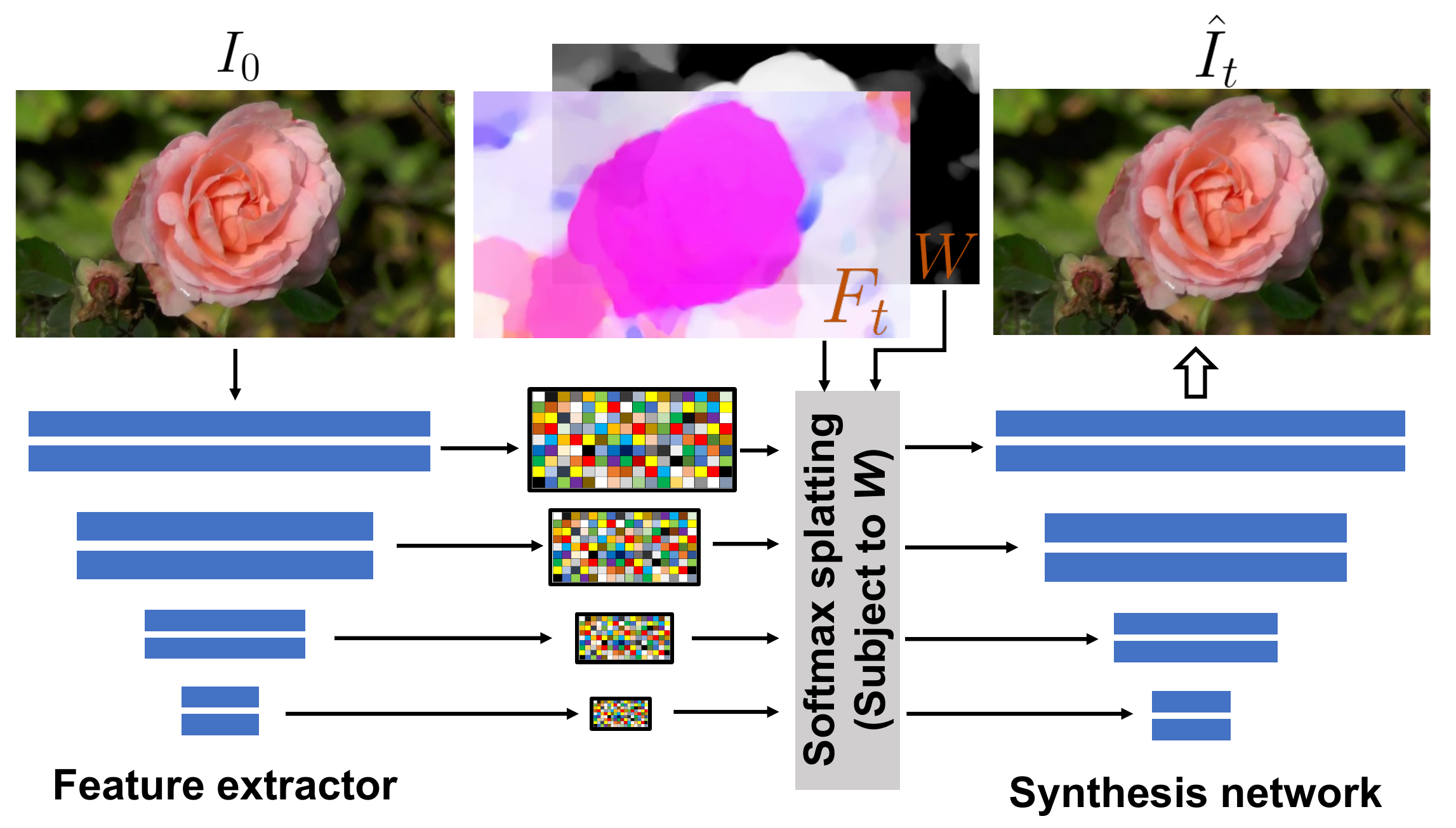} 
  \caption{\textbf{Rendering module.} We fill in missing content and refine the warped input image using a deep image-based rendering module, where multi-scale features are extracted from the input image $I_0$. Softmax splatting is then applied over the features with a motion field $\FLOW_t$ from time $0$ to $t$ (subject to the weights $W$). The warped features are fed to an image synthesis network to produce the rendered image $\hat{I}_t$. }
\label{fig:overview_2}
\end{figure}

We now describe how we take a spectral volume $\SMT$ predicted for a given input image $I_0$ and render a future frame $\hat{I}_t$ at time $t$. 
We first derive a motion texture in the time domain using the inverse temporal FFT applied at each pixel $\MT (\pix) = \text{FFT}^{-1} (\SMT (\pix))$. 
To produce a future frame $\hat{I}_t$, we adopt a deep image-based rendering technique and perform splatting with the predicted motion field $\FLOW_t$ to forward-warp the encoded $I_0$, as shown in Fig.~\ref{fig:overview_2}. 
Since forward warping can lead to holes, and multiple source pixels can map to the same output 2D location, we adopt the feature pyramid softmax splatting strategy proposed in prior work on frame interpolation~\cite{niklaus2020softmax}.

Specifically, we encode $I_0$ through a feature extractor network to produce a multi-scale feature map. %
For each individual feature map at scale $j$, we resize and scale the predicted 2D motion field $\FLOW_t$ according to the resolution. As in Davis \etal~\cite{davis2015image}, we use predicted flow magnitude as a proxy for depth to determine the contributing weight of each source pixel mapped to its destination location. In particular, we compute a per-pixel weight, $W (\pix)= \frac{1}{T} \sum_{t} || \FLOW_t (\pix)||_2$ as the average magnitude of the predicted motion texture. 
In other words, we assume large motions correspond to moving foreground objects, and small or zero motions correspond to background. 
We use motion-derived weights instead of learnable ones as \cite{holynski2021animating} because we observe that in the single-view case, learnable weights are not effective for addressing disocclusion ambiguities.

With the motion field $\FLOW_t$ and weights $W$, we apply softmax splatting to warp the feature map at each scale to produce a warped feature.
The warped features are then injected into the corresponding blocks of an image synthesis decoder to produce a final rendered image $\hat{I}_t$. 

We jointly train the feature extractor and synthesis networks with start and target frames $(I_0, I_t)$ randomly sampled from real videos, using the estimated flow field from $I_0$ to $I_t$ to warp encoded features from $I_0$, and supervising predictions $\hat{I}_t$ against $I_t$ with a VGG perceptual loss~\cite{johnson2016perceptual}.

\section{Applications} \label{sec:app}

\paragraph{Image-to-video.} 
\label{sec:image2video}
Our system enables the animation of a single still picture by first predicting a motion spectral volume from the input image and generating an animation by applying our image-based rendering module to the motion texture transformed from the spectral volume. Since we explicitly model scene motion, this allows us to produce slow-motion videos by linearly interpolating the motion texture, or to magnify (or minify) animated motions by adjusting the amplitude of the predicted spectral volume coefficients.

\paragraph{Seamless looping.}
\label{sec:looping}
Many applications require videos that loop seamlessly, 
where there is no discontinuity 
between the start and end of the video. 
Unfortunately, it is hard to find a large collection of seamlessly looping videos for training.
Instead, we devise a method to use our motion diffusion model, trained on regular non-looping video clips, to produce seamless looping video. 
Inspired by recent work on guidance for image editing~\cite{epstein2023diffusion,bansal2023universal}, our method is a \emph{motion self-guidance} technique that guides the motion denoising sampling processing using explicit looping constraints. 
In particular, at each iterative denoising step during inference, we incorporate an additional motion guidance signal alongside standard classifier-free guidance~\cite{ho2022classifier}, where we enforce each pixel's position and velocity at the start and end frames to be as similar as possible:
\begin{multline}
    \hat{\epsilon}^n = (1 + w) \epsilon_{\theta}(z^n; n, c) - w \epsilon_{\theta} (z^n; n, \emptyset) 
    +  u \sigma^n \nabla_{z^n} \mathcal{L}^n_g  \\
    \mathcal{L}^n_g  = ||\FLOW^{n}_{T} - \FLOW^{n}_{1}||_1 + ||\nabla \FLOW^{n}_{T} - \nabla \FLOW^{n}_{1}||_1
\end{multline}
where $\FLOW^{n}_{t}$ is the predicted 2D displacement field at time $t$ and denoising step $n$. $w$ is the classifier-free guidance weight, and $u$ is the motion self-guidance weight. 
In the supplemental video, we apply a baseline appearance-based looping algorithm~\cite{liao2015fast} to generate a looping video from our non-looping output, and show that our motion self-guidance technique produces seamless looping videos with less distortion and fewer artifacts.

\paragraph{Interactive dynamics from a single image.} \label{sec:interactive}
Davis \etal~\cite{davis2015image} show that
the spectral volume, evaluated at certain resonant frequencies, can approximate
an \emph{image-space modal basis} that is a projection of the vibration modes of the underlying scene (or, more generally, captures spatial and temporal correlations in oscillatory dynamics), and can be used to simulate the object's response to a user-defined force. %
We adopt this modal analysis method~\cite{davis2015image, petitjean2023modalnerf}, 
allowing us to write the image-space 2D motion displacement field for the object's physical response as a weighted sum of motion spectrum coefficients $S_{f_j}$ modulated by the state of complex modal coordinates $\mathbf{q}_{f_j}(t)$ at each simulated time step $t$:
\begin{align}
    \FLOW_t ({\pix}) = \sum_{f_j} S_{f_j}(\pix) \mathbf{q}_{f_j} (t) 
\end{align}
We simulate the state of the modal coordinates $\mathbf{q}_{f_j}(t)$ via an explicit Euler method applied to the equations of motion for a decoupled mass-spring-damper system represented in modal space~\cite{davis2015image, davis2016visual, petitjean2023modalnerf}. We refer readers to supplementary material and original work for a full derivation. Note that our method produces an interactive scene from a \emph{single picture}, whereas these prior methods required a video as input.

\section{Experiments}

\newcommand{\tablespace}{\,\,\,\,}
\newcommand{\halftablespace}{\,}
\setlength{\tabcolsep}{1.5pt}
\begin{table}[t]
\begin{center}
\footnotesize
\begin{tabular}{l cc @{\hskip 1em} cccc}
\toprule
& \multicolumn{2}{c@{\hskip 1em}}{Image Synthesis} & \multicolumn{4}{c@{\hskip 0.5em}}{Video Synthesis}
\\
Method & FID & KID & $\text{FVD}$ & $\text{FVD}_{32}$ & $\text{DTFVD}$ & $\text{DTFVD}_{32}$
\\ 
\midrule
TATS~\cite{ge2022long} & 65.8 & 1.67 & 265.6 & 419.6 & 22.6 & 40.7 \\ 
Stochastic I2V~\cite{Dorkenwald_2021_CVPR} & 68.3 & 3.12 & 253.5 & 320.9 & 16.7 & 41.7  \\
MCVD~\cite{voleti2022MCVD} & 63.4 & 2.97 & 208.6 & 270.4 & 19.5 & 53.9 \\
LFDM~\cite{ni2023conditional} & 47.6 & 1.70 & 187.5 & 254.3 & 13.0 & 45.6 \\
DMVFN~\cite{Hu2023ADM} & 37.9 & 1.09 & 206.5 & 316.3 & 11.2 & 54.5 \\ 
Endo~\etal~\cite{endo2019animatinglandscape} & 10.4 & 0.19 & 166.0 & 231.6 & 5.35 & 65.1 \\
Holynski~\etal~\cite{holynski2021animating} & 11.2 & 0.20 & 179.0 & 253.7 & 7.23 & 46.8 \\
Ours & \textbf{4.03} & \textbf{0.08} & \textbf{47.1} & \textbf{62.9} & \textbf{2.53} & \textbf{6.75} \\
\bottomrule
\end{tabular}
\caption{\textbf{Quantitative comparisons on the test set.} We report both image synthesis and video synthesis quality. Here, KID is scaled by 100. Lower is better for all error.
See Sec.~\ref{sec:quantitative} for descriptions of baselines and error metrics. } \label{table:test_number} 
\end{center} 
\end{table}

\begin{figure*}[tb]
    \centering
    \setlength{\tabcolsep}{0.02cm}
    \begin{tabular}{cccccccc}
        \includegraphics[width=0.29\columnwidth]{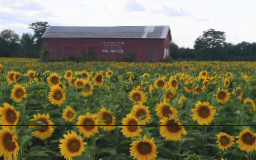} & 
        \includegraphics[width=0.29\columnwidth]{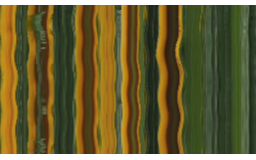} & 
        \includegraphics[width=0.29\columnwidth]{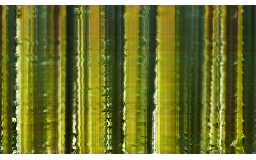} & 
        \includegraphics[width=0.29\columnwidth]{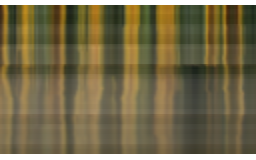} & 
        \includegraphics[width=0.29\columnwidth]{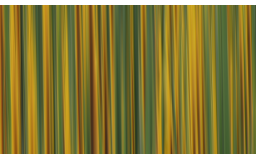} &
        \includegraphics[width=0.29\columnwidth]{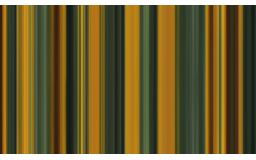} &
        \includegraphics[width=0.29\columnwidth]{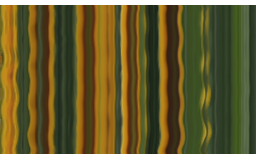} 
        \\
        \includegraphics[width=0.29\columnwidth]{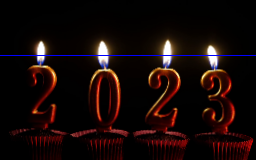} & 
        \includegraphics[width=0.29\columnwidth]{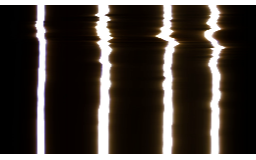} & 
        \includegraphics[width=0.29\columnwidth]{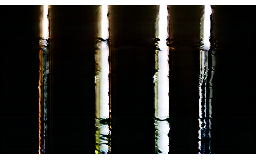} & 
        \includegraphics[width=0.29\columnwidth]{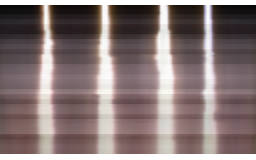} & 
        \includegraphics[width=0.29\columnwidth]{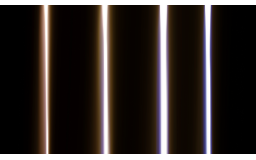} &
        \includegraphics[width=0.29\columnwidth]{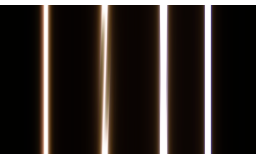} &
        \includegraphics[width=0.29\columnwidth]{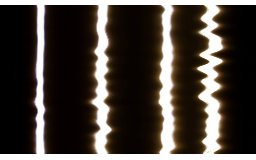} 
        \\
        \includegraphics[width=0.29\columnwidth]{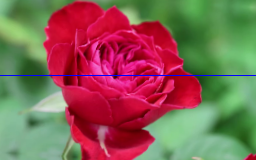} & 
        \includegraphics[width=0.29\columnwidth]{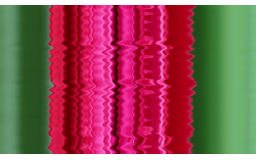} &
        \includegraphics[width=0.29\columnwidth]{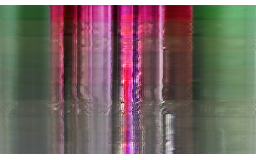} & 
        \includegraphics[width=0.29\columnwidth]{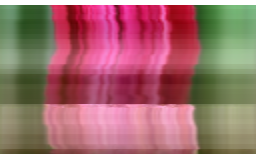} &
        \includegraphics[width=0.29\columnwidth]{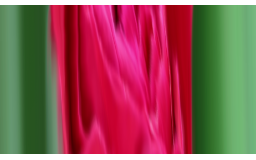} &
        \includegraphics[width=0.29\columnwidth]{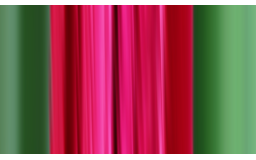} &
        \includegraphics[width=0.29\columnwidth]{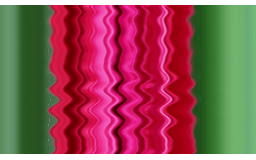}
        \\
        \includegraphics[width=0.29\columnwidth]{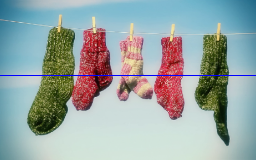} & 
        \includegraphics[width=0.29\columnwidth]{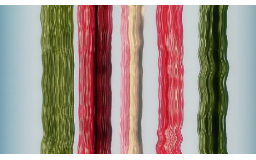} & 
        \includegraphics[width=0.29\columnwidth]{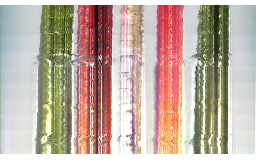} &
        \includegraphics[width=0.29\columnwidth]{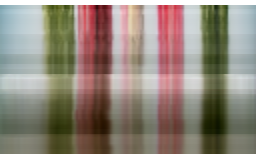} &
        \includegraphics[width=0.29\columnwidth]{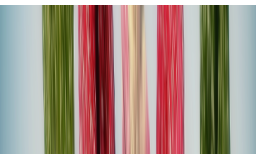} &
        \includegraphics[width=0.29\columnwidth]{figures/epi/3717776_1_epi_endo.png} &
        \includegraphics[width=0.29\columnwidth]{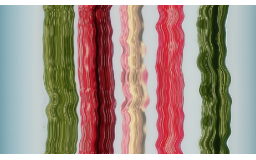}
        \\
        \vphantom{I} \footnotesize{Input image} &
        \vphantom{I} \footnotesize{Reference} &
        \vphantom{I} \footnotesize{Stochastic-I2V}~\cite{Dorkenwald_2021_CVPR} &
        \vphantom{I} \footnotesize{MCVD}~\cite{voleti2022MCVD} &
        \vphantom{I} \footnotesize{Endo~\etal}~\cite{endo2019animatinglandscape} &
        \vphantom{I} \footnotesize{Holynski~\etal}~\cite{holynski2021animating} &
        \vphantom{I} \footnotesize{Ours} 
    \end{tabular}
    \vspace{-0.2cm}
    \caption{\textbf{$X$-$t$ slices of videos generated by different approaches.} From left to right: input image and corresponding $X$-$t$ video slices from the ground truth video, from videos generated by three baselines~\cite{Dorkenwald_2021_CVPR, voleti2022MCVD, endo2019animatinglandscape, holynski2021animating}, and finally videos generated by our approach.}
    \label{fig:viz_comp_1}
\end{figure*}

\paragraph{Implementation details.}
We use an LDM~\cite{rombach2022high} as the backbone for predicting spectral volumes, for which we use a VAE with a continuous latent space of dimension $4$. We train the VAE with an $L_1$ reconstruction loss, a multi-scale gradient consistency loss~\cite{li2018megadepth, li2019learning, li2023dynibar}, and a KL-divergence loss with respective weights of $1, 0.2, 10^{-6}$. We train the same 2D U-Net used in the original LDM work to perform iterative denosing with a simple MSE loss~\cite{ho2020denoising}, and adopt the attention layers from~\cite{he2022latent} for frequency-coordinated denoising. 
For quantitative evaluation, we train both VAE and LDM on images of size $256\times160$ from scratch for fair comparisons, and it takes around 6 days to converge using 16 Nvidia A100 GPUs. 
For our main quantitative and qualitative results, we run the motion diffusion model with DDIM~\cite{song2020denoising} for 250 steps.
We also show generated videos of up to a resolution of $512\times288$, created by fine-tuning a pre-trained image inpainting LDM model~\cite{rombach2022high} on our dataset.

We adopt ResNet-34~\cite{he2016deep} for the feature extractor in our IBR module.
Our image synthesis network is based on an architecture for conditional image inpainting~\cite{zhao2021comodgan, li2022infinitenature}. 
Our rendering module runs in real-time at 25FPS on a Nvidia V100 GPU during inference. We adopt universal guidance~\cite{bansal2023universal} to produce seamless looping videos, where we set weights $w=1.75, u=200$, and use 500 DDIM steps with 2 self-recurrence iterations.

\paragraph{Data.}
We collect and process a set of 3,015 videos of natural scenes exhibiting oscillatory motions from online sources 
our own captures. We withhold 10\% of 
videos for testing and use the remainder for training. 
To extract ground truth motion trajectories, 
we apply a coarse-to-fine 
flow method~\cite{brox2004high, liu2009beyond} between each selected starting image and every future frame of 
the video. As training data, we take every 10th video frame as input images and derive the corresponding ground truth spectral volumes using the computed motion trajectories across the following 149 frames. 
In total, our data consists of over 150K image-motion pairs.

\begin{figure}[tb]
    \centering
    \includegraphics[width=\columnwidth]{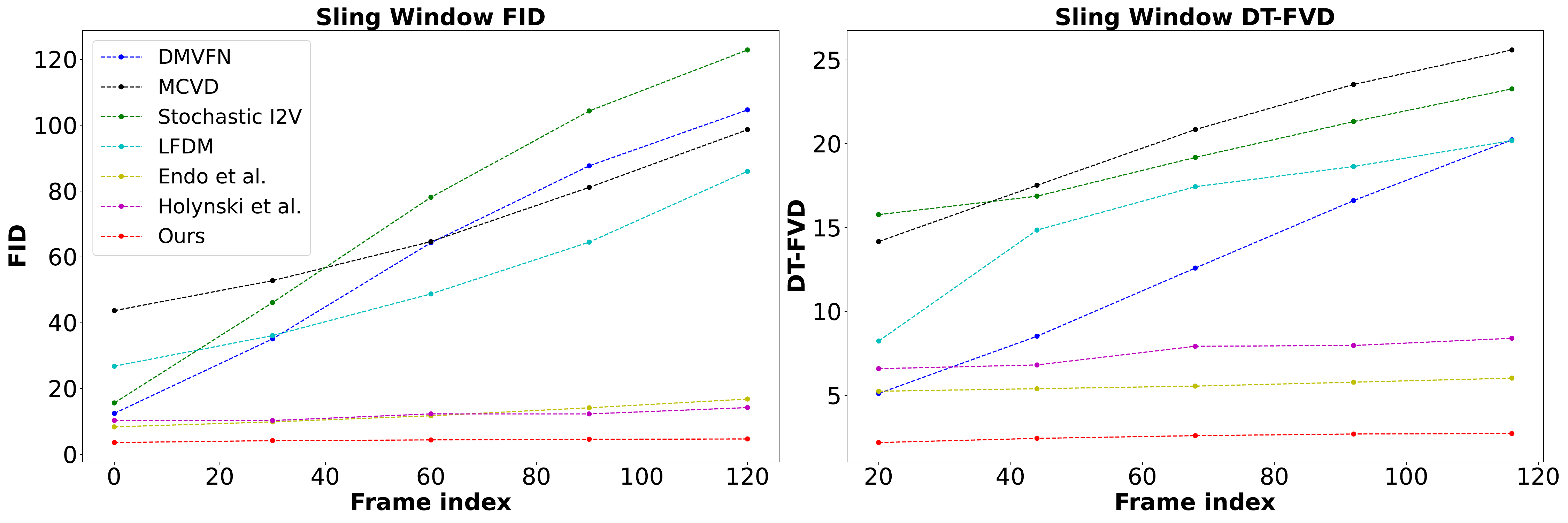}
    \vspace{-0.5cm}
    \caption{\textbf{Sliding window FID and DTFVD.} We show sliding window FID with window size 30 frames, and DTFVD with size 16 frames, for videos generated by different methods.}
    \label{fig:fid_fvd}
\end{figure}

\paragraph{Baselines.}
We compare our approach to recent single-image animation and video prediction methods. Endo~\etal~\cite{endo2019animatinglandscape} and DMVFN~\cite{Hu2023ADM} predict instantaneous 2D motion fields and render future frames auto-regressively. Holynski~\etal~\cite{holynski2021animating} instead simulates motion through a single static Eulerian motion description. Other recent work such as Stochastic Image-to-Video (Stochastic-I2V)~\cite{Dorkenwald_2021_CVPR}, TATS~\cite{ge2022long}, and MCVD~\cite{voleti2022MCVD} adopt VAEs, transformers, or diffusion models to directly predict raw video frames; LFDM~\cite{ni2023conditional} generates future frames by predicting flow volumes and warping latents in a diffusion model.
We train all the above methods on our data using their respective open-source implementations.\footnote{We use the open-source implementation of \cite{holynski2021animating} from Fan~\etal~\cite{fan2022SLR}.}

We evaluate the quality of the videos generated by our approach and by prior baselines in two ways.
First, we evaluate the quality of individual synthesized frames using metrics designed for image synthesis tasks. 
We adopt the Fréchet Inception Distance (FID)~\cite{heusel2017gans} and Kernel Inception Distance (KID)~\cite{binkowski2018demystifying} to measure the average distance between the distributions of generated frames and ground truth frames. 

Second, to evaluate the quality and temporal coherence of synthesized videos, we adopt the Fréchet Video Distance~\cite{unterthiner2018towards} with window size 16 (FVD) and 32 ($\text{FVD}_{32}$), based on an I3D model~\cite{carreira2017quo} trained on the Human Kinetics datasets~\cite{kay2017kinetics}. To more faithfully reflect synthesis quality for the natural oscillation motions we seek to generate, we also adopt the Dynamic Texture Frechet Video Distance~\cite{Dorkenwald_2021_CVPR}, which measures distance from videos with window size 16 (DTFVD) and size 32 ($\text{DTFVD}_{32}$), using a I3D model trained on the Dynamic Textures Database~\cite{hadji2018new}, a dataset consisting primarily of natural motion textures. 

We further use a sliding window FID of window size 30 frames, and a sliding window DTFVD with window size 16 frames, as in~\cite{liu2021infinite, li2022infinitenature}, to measure how generated video quality degrades over time.
For all methods, we evaluate metrics at $256\times128$ resolution by center-cropping. %

\renewcommand{\tablespace}{\,\,\,\,}
\renewcommand{\halftablespace}{\,}
\setlength{\tabcolsep}{1.5pt}
\begin{table}[t]
\begin{center}
\footnotesize
\begin{tabular}{l cc @{\hskip 1em} cccc}
\toprule
& \multicolumn{2}{c@{\hskip 1em}}{Image Synthesis} & \multicolumn{4}{c@{\hskip 0.5em}}{Video Synthesis}
\\
Method & FID & KID & $\text{FVD}$ & $\text{FVD}_{32}$ & $\text{DTFVD}$ & $\text{DTFVD}_{32}$
\\
\midrule
Repeat $I_0$ & - & - & 237.5 & 316.7 & 5.30 & 45.6 \\
\midrule
$K=4$ & 3.92 & 0.07 & 60.3 & 78.4 & 3.12 & 8.59 \\
$K=8$ & 3.95 & 0.07 & 52.1 & 68.7 & 2.71 & 7.37 \\
$K=24$ & 4.09 & 0.08 & 48.2 & 65.1 & 2.50 & 6.94 \\
\midrule
w/o adaptive norm. & 4.53 & 0.09 & 62.7 & 80.1 & 3.16 & 8.19 \\
Independent pred. & 4.00 & 0.08 & 52.5 & 71.3 & 2.70 & 7.40  \\
Volume pred. & 4.74 & 0.09 & 53.7 & 71.1 & 2.83 & 7.79  \\
\midrule
Baseline splat~\cite{holynski2021animating} & 4.25 & 0.09 & 49.5 & 66.8 & 2.83 & 7.27 \\
\midrule
Full ($K=16$)  & {4.03} & {0.08} & {47.1} & {62.9} & {2.53} & {6.75} \\
\bottomrule
\end{tabular}
\vspace{-0.2cm}
\caption{\textbf{Ablation study.} Sec.~\ref{sec:ablation} describes each configuration.} \label{table:ablation}  
\end{center}
\end{table}

\subsection{Quantitative results}  \label{sec:quantitative}

Table~\ref{table:test_number} shows quantitative comparisons between our approach and baselines on our test set.
Our approach significantly outperforms prior single-image animation baselines in terms of both image and video synthesis quality. 
Specifically, our much lower FVD and DT-FVD distances suggest that the videos generated by our approach are more realistic and more temporally coherent.
Further, Fig.~\ref{fig:fid_fvd} shows the sliding window FID and sliding window DT-FVD distances of generated videos from different methods. Thanks to the global spectral volume representation, videos generated by our approach do not suffer from degradation over time. %

\subsection{Qualitative results}  \label{sec:qualitative}

We visualize qualitative comparisons between videos as spatio-temporal $X$-$t$ slices of the generated videos, a standard way of visualizing small motions in a video~\cite{wadhwa2013phase}. 
As shown in Fig.~\ref{fig:viz_comp_1}, our generated video dynamics more strongly resemble the motion patterns observed in the corresponding real reference videos (second column), compared to other methods. Baselines such as Stochastic I2V~\cite{Dorkenwald_2021_CVPR} and MCVD~\cite{voleti2022MCVD} fail to model both appearance and motion realistically over time.  Endo~\etal~\cite{endo2019animatinglandscape} and Holynski~\etal~\cite{holynski2021animating} produce video frames with fewer artifacts but exhibits over-smooth or non-oscillatory motion over time. 
We refer readers to supplementary material to assess the quality of generated video frames and estimated motion across different methods.

\begin{figure}[tb]
    \centering
    \setlength{\tabcolsep}{0.01cm}
    \begin{tabular}{cccc}
        \includegraphics[width=0.25\columnwidth]{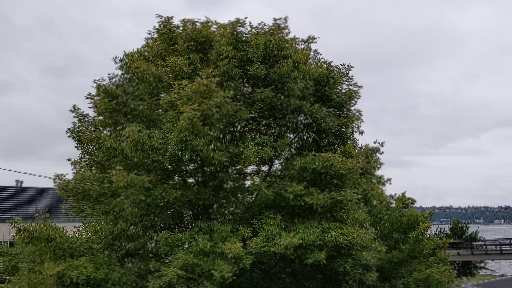} &
        \includegraphics[width=0.25\columnwidth]{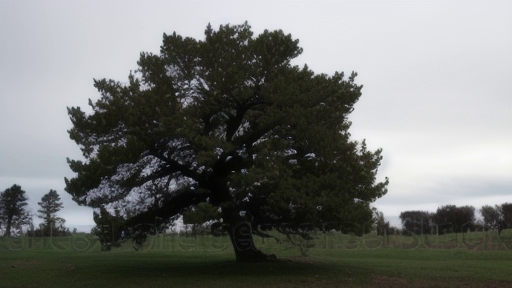} &
        \includegraphics[width=0.25\columnwidth]{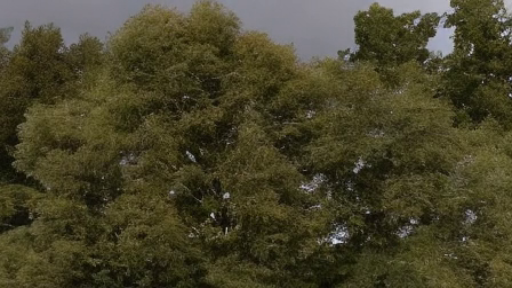} &
        \includegraphics[width=0.25\columnwidth]{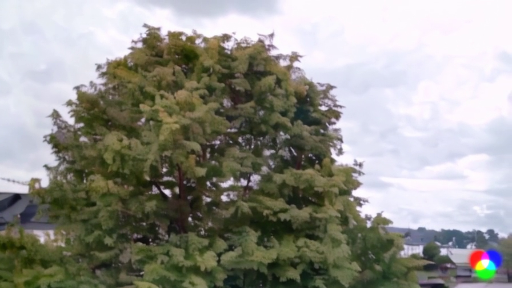}
        \\
        \vphantom{I} Input &
        \vphantom{I} AnimateDiff &
        \vphantom{I} ModelScope &
        \vphantom{I} GEN-2
    \end{tabular}
    \vspace{-0.3cm}
    \caption{We show generated future frames from three recent large video diffusion models~\cite{guo2023animatediff, wang2023videocomposer, esser2023structure}.} %
    \label{fig:llm}
\end{figure}

\subsection{Ablation study}\label{sec:ablation}
We conduct an ablation study to validate the major design choices in our motion prediction and rendering modules, comparing our full configuration with different variants. 
Specifically, we evaluate results using different numbers of frequency bands $K=4, 8, 16, 24$. We observe that increasing the number of frequency bands improves video prediction quality, but the improvement is marginal with more than 16 frequencies. 
Next, we remove adaptive frequency normalization from the ground truth spectral volumes, and instead just scale them based on input image width and height (\emph{w/o adaptive norm.}). 
Additionally, we remove the frequency coordinated-denoising module (\emph{Independent pred.}), or replace it with a simpler DM where a tensor volume of $4K$ channel spectral volumes are predicted jointly via a single 2D U-net diffusion model (\emph{Volume pred.}). Finally, we compare results where we use a baseline rendering method that applies softmax splatting over single-scale features subject to learnable weights as used by~\cite{holynski2021animating} (\emph{Baseline splat}). We also add a baseline where the generated video is a volume by repeating input image $N$ times (Repeat $I_0$).
From Table~\ref{table:ablation}, we observe that all simpler or alternative configurations lead to worse performance compared with our full model.

\subsection{Comparing to large video models}  \label{sec:user_study}
We further perform a user study and compare our generated animations with ones from recent large video diffusion models: AnimateDiff~\cite{guo2023animatediff}, ModelScope~\cite{wang2023videocomposer} and Gen-2~\cite{esser2023structure}, which predict video volumes directly. On a randomly selected 30 videos from the test set, we ask users ``which video is more realistic?''. Users report a 80.9\% preference for our approach over others. 
Moreover, as shown in Fig.~\ref{fig:llm}, we observe that the generated videos from these baselines are either unable to adhere to the input image content, or exhibit gradual color drift and distortion over time. We refer readers to the supplementary material for full comparisons.

\section{Discussion and conclusion}

\begin{figure}[tb]
    \centering
    \setlength{\tabcolsep}{0.01cm}
    \begin{tabular}{cccc}
        \includegraphics[width=0.25\columnwidth]{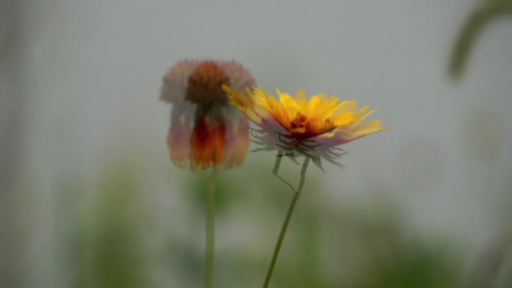} &
        \includegraphics[width=0.25\columnwidth]{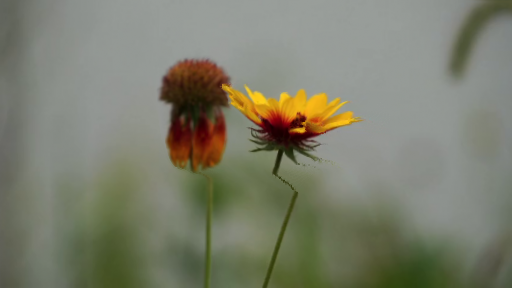} &
        \includegraphics[width=0.25\columnwidth]{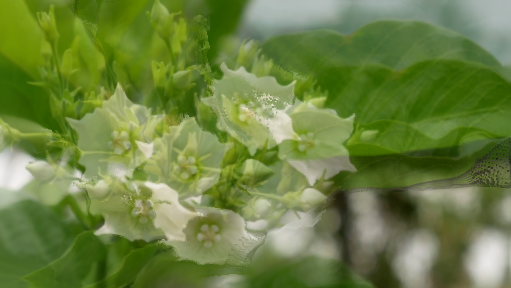} &
        \includegraphics[width=0.25\columnwidth]{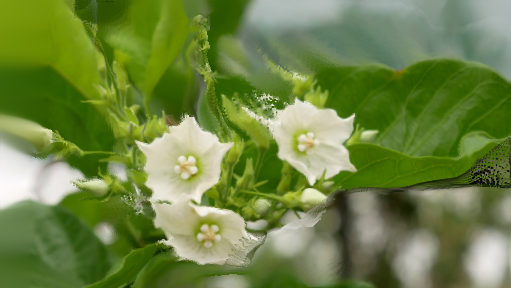} \\
    \end{tabular}
    \vspace{-0.3cm}
    \caption{\textbf{Limitations.} We show examples of rendered future frames (even), and overlay of input and rendered images (odd). Our method can produce artifacts in regions of thin objects or large motions, and regions requiring filling large amount of new contents.} \label{fig:limitations}
\end{figure}

\paragraph{Limitations.} 
Since our approach only predicts lower frequencies of a spectral volume, it can fail to model non-oscillating motions or high-frequency vibrations---this may be resolved by using learned motion bases.
Furthermore, the quality of generated videos relies on the quality of underlying motion trajectories, which may degrade in scenes with thin moving objects or objects with large displacements. Even if correct, motions that require generating large amounts of new unseen content may also cause degradations (Fig.~\ref{fig:limitations}). 

\paragraph{Conclusion.}
We present a new approach for modeling natural oscillation dynamics from a single still picture. Our image-space motion prior is represented with spectral volumes, a frequency representation of per-pixel motion trajectories, which we find to be efficient and effective for prediction with diffusion models, and which we learn from collections of real world videos. 
Spectral volumes are predicted using frequency-coordinated latent diffusion model and are used to animate future video frames through an image-based rendering module. 
We show that our approach produces photo-realistic animations from a single picture and significantly outperforms prior baselines, and that it can enable several downstream applications such as creating seamlessly looping or interactive image dynamics.

\paragraph{Acknowledgements.}
We thank \newtext{Abe Davis}, Rick Szeliski, Andrew Liu, Boyang Deng, Qianqian Wang, Xuan Luo, and Lucy Chai for fruitful discussions and helpful comments.

{\small
\bibliographystyle{ieee_fullname}
\bibliography{refs}
}

\end{document}